%% file: main.tex
\documentclass[10pt,twocolumn,letterpaper]{article}
\usepackage{arxiv}              

\input{macros}

\usepackage[pagebackref,breaklinks,colorlinks,allcolors=blue]{hyperref}

\begin{document}
\title{\method: Scene-Aware 3D Human Motion Synthesis by Planning with Geometry-Grounded Tokens}

\author{
Anindita Ghosh$^{1,2,3}$\hspace{2.3em}
Vladislav Golyanik$^{2,3}$\hspace{2.3em} 
Taku Komura$^{4}$\hspace{2.3em}
Philipp Slusallek$^{1,3}$\vspace{0.3em}\\
Christian Theobalt$^{2,3}$\hspace{1.3em}
Rishabh Dabral$^{2,3}$\vspace{0.3em}\\
\textit{$^1$DFKI}\vspace{0.3em} \hspace{1.3em}
\textit{$^2$MPI for Informatics}\vspace{0.3em} \hspace{1.3em}
\textit{$^3$Saarland Informatics Campus}\vspace{0.3em} \hspace{1.3em}
\textit{$^4$University of Hong Kong}\\
\hspace{1.3em}
\texttt{\href{https://anindita127.github.io/SceMoS/}{https://anindita127.github.io/SceMoS/}}
}
\maketitle
\input{sec/0_abstract}    
\input{sec/1_introduction}

\input{sec/2_relatedwork}
\input{sec/3_method}
\input{sec/4_experiments}
\input{sec/5_results}
\input{sec/6_conclusion}
{
    \small
    \bibliographystyle{ieeenat_fullname}
    \bibliography{main}
}
\clearpage
\appendix

\input{sec/7_appendix}
\end{document}

%% file: macros.tex
\usepackage{amsmath}
\usepackage{amsfonts}
\usepackage{amsthm}
\usepackage{array}
\usepackage{balance}
\usepackage{bbm}
\usepackage{booktabs}

\usepackage{color, colortbl}
\usepackage{enumitem}
\usepackage{float}
\usepackage[symbol]{footmisc}
\usepackage{mathtools}
\usepackage{mathrsfs}
\usepackage{multirow}
\usepackage{pifont}
\usepackage{setspace}
\usepackage{subcaption}
\usepackage[normalem]{ulem}
\usepackage{xcolor}
\usepackage{xfrac}
\usepackage[ruled]{algorithm2e} 
\usepackage{algpseudocode}

\SetAlFnt{\small}
\SetAlCapFnt{\small}
\SetAlCapNameFnt{\small}
\SetAlCapHSkip{0pt}

\theoremstyle{definition}

\newcolumntype{L}[1]{>{\raggedright\let\newline\\\arraybackslash\hspace{0pt}}m{#1}}
\newcolumntype{C}[1]{>{\centering\let\newline\\\arraybackslash\hspace{0pt}}m{#1}}
\newcolumntype{R}[1]{>{\raggedleft\let\newline\\\arraybackslash\hspace{0pt}}m{#1}}

\setlist[itemize]{noitemsep, topsep=0pt}
\setlist[enumerate]{noitemsep, topsep=0pt}

\DeclareUnicodeCharacter{3000}{ }
\DeclareUnicodeCharacter{2212}{-}

\DeclareMathOperator*{\argmax}{arg\,max}

\newcommand{\bracks}[1]{\left[#1\right]}

\newcommand{\norm}[1]{\left\Vert#1\right\Vert}

\newcommand{\bulletitem}{\item[$\bullet$]}

\newcommand{\method}{SceMoS}
\definecolor{Gray}{gray}{0.9}
\definecolor{modification_color}{rgb}{1.0, 0.0, 0.0}

%% file: sec/0_abstract.tex
\begin{abstract}
   Synthesizing text-driven 3D human motion within realistic scenes requires learning both semantic intent (``walk to the couch'') and physical feasibility (e.g.,~avoiding collisions). 
   Current methods use generative frameworks that simultaneously learn high-level planning and low-level contact reasoning, and rely on computationally expensive 3D scene data such as point clouds or voxel occupancy grids.
   We propose \method, a scene-aware motion synthesis framework that shows that structured 2D scene representations can serve as 
   a powerful alternative to full 3D supervision in physically grounded motion synthesis. 
   \method~ disentangles global planning from local execution using lightweight 2D cues and relying on  
   (1) a text-conditioned autoregressive global motion planner that operates on a bird’s-eye-view (BEV) image rendered from an elevated corner of the scene, encoded with DINOv2 features, as the scene representation, and
   (2) a geometry-grounded motion tokenizer trained via a conditional VQ-VAE, that uses 2D local scene heightmap, thus embedding surface physics directly into a discrete vocabulary.
   This 2D factorization reaches an efficiency-fidelity trade-off: BEV semantics capture spatial layout and affordance for global reasoning, while local heightmaps enforce fine-grained physical adherence without full 3D volumetric reasoning. 
   \method~achieves state-of-the-art motion realism and contact accuracy on the TRUMANS benchmark, reducing the number of trainable parameters for scene encoding by over $50\%$, showing that 2D scene cues can effectively ground 3D human-scene interaction.
\end{abstract}

\begin{figure}[t]
  \centering
   \includegraphics[width=\linewidth]{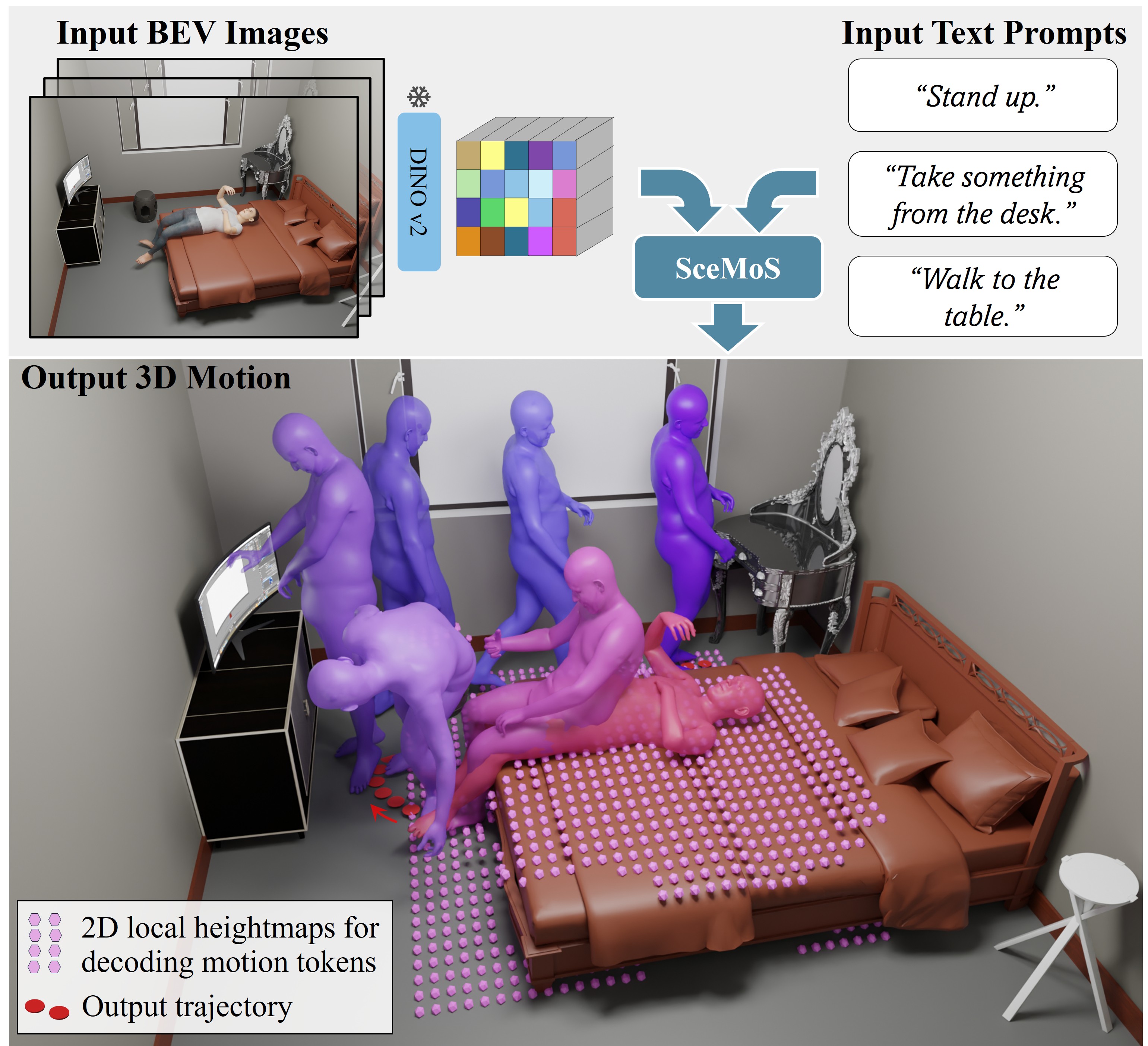}
   \caption{The introduced scene-aware 3D human motion synthesis framework, \textbf{\method}, uses 2D scene cues and text instructions to generate physically consistent and realistic 3D motions. We use a bird's-eye-view (BEV) image rendered from an elevated corner of the input scene, and extract DINOv2 features for high-level semantic planning. For fine-grained contact reasoning, we use the local 2D heightmap of the scene around the root of the person's initial pose. 
   }
   \label{fig:teaser}
   \vspace{-15pt}
\end{figure}

%% file: sec/1_introduction.tex
\section{Introduction} \label{sec:introduction}
The field of generative 3D human motion synthesis has seen remarkable progress ~\cite{zhang2022motiondiffuse, Ghosh_2021_ICCV,li2025simmotionedit, tevet2023MDM, dabral2022mofusion}
 in animation, virtual reality, robotics, and embodied AI~\cite{hanser2009scenemaker, yoon2019robots, egges2007presence}  in recent years. 
As research has advanced, introducing scene-awareness to text-guided motion synthesis has emerged as a critical next step since human actions are inherently shaped by their surrounding environment. Our work addresses the challenge of generating scene-aware human motions in the paradigm of human-scene interactions (HSI), driven by textual intent and grounded in the spatial context of the scene.
\par
Achieving this goal in a data-driven framework raises a fundamental question: \textit{How should the 3D scene be represented relative to the character?} Prior efforts have explored a range of scene representations, from coarse layouts such as CAD primitives~\cite{somani2015object} to dense 3D geometry such as voxel grids~\cite{maturana2015voxnet}, point clouds~\cite{qi2017pointnet}, and signed distance fields~\cite{oleynikova2016signed}.
Meanwhile, human-object interaction studies have adopted object-centric representations such as Basis Point Sets (BPS)~\cite{ghosh2022imos, GRAB:2020}. 
\par
However, a tradeoff persists: \textit{scene-level} representations are either bulky or lack the detail required for fine-grained contact reasoning, whereas \textit{object-level} or dense 3D formats are computationally expensive and difficult to scale to full scenes.
For instance, the cubic memory complexity of voxel grids and the unstructured nature of point clouds require heavy 3D backbones such as volumetric CNNs~\cite{teng2022structured} or transformer-based architectures~\cite{yi2024tesmo},
simply to interpret the scene, thus introducing redundant spatial detail for tasks dominated by near-surface interactions. 
Consequently, current methods must simultaneously learn to perceive complex geometry, plan spatial intent, and execute fine-grained motion within a single, entangled animation process~\cite{neural_state_machines, starke2022deepphase}.
This trade-off is particularly prohibitive for synthesizing human motion in realistic scenes, where scene assets are noisy and unlabeled.
Here, the motions must remain semantically consistent with textual intent (\textit{e.g.}, ``sit on the couch'') while also being physically compliant with the surrounding geometry (\textit{e.g.}, not walking through furniture to reach the couch).
This dual requirement of semantic and physical consistency presents a fundamental challenge, and current state-of-the-art methods~\cite{Zhao2023dimos, wang2022humanise, wang2021scene} 
face a persistent conflict between generalization and efficiency. 
\par
In this work, we revisit the question of scene representations and explore how structured 2D scene representations can provide sufficient cues for physically grounded motion needed for HSI.
Human interaction with the environment is largely guided by two complementary types of 2D information, \textit{i.e.},~global layout cues that inform spatial reasoning, and local geometric cues that guide physical contact. 
Inspired by this, our scene-aware motion synthesis framework, \method, explicitly disentangles global planning from local execution. 
We employ lightweight 2D scene representations at both levels. 
For the text-conditioned motion planning, we use a bird’s-eye-view (BEV) image rendered from an elevated corner of the scene, with the camera pointed toward the target character from the elevated corner to ensure minimal occlusions. 
This allows us to leverage the strong semantic representation abilities of vision foundational models like DINOv2~\cite{oquab2023dinov2} for high-level 3D semantic planning.
For fine-grained interaction reasoning, we compute the local 2D heightmap around the  character from their previous 3D pose, thereby providing a more precise, but local geometric representation. 
\cref{fig:teaser} illustrates the overview of our approach. 
Together, these cues retain essential spatial and geometric information while avoiding the redundancy and cost of dense 3D volumetric reasoning.
\par
In summary, our work makes the following contributions toward efficient, physically grounded HSI synthesis:
\begin{itemize}
    \item A \textbf{lightweight two-stage framework for text-driven HSI generation} that explicitly disentangles global motion planning from local physical execution. This separation enables efficient reasoning over complex scenes without relying on dense 3D volumetric inputs.
    
    \item A \textbf{geometry-grounded motion vocabulary} learned via a conditional VQ-VAE, which explicitly conditions the motion decoder on local scene 2D heightmaps to ensure physical plausibility.
    
    \item A \textbf{global motion planner conditioned on compact 2D scene cues}, combining DINOv2 features from a single BEV image with text embeddings. We demonstrate that this 2D representation captures sufficient spatial and semantic context for long-horizon, goal-directed motion synthesis, significantly lowering computational cost for scene encoding.
\end{itemize}

\method~achieves state-of-the-art realism and contact accuracy on the TRUMANS dataset~\cite{jiang2024scaling}, being as efficient as the diffusion or point-cloud-based approaches,
while requiring an order of magnitude fewer trainable scene parameters (${\sim}4M$ vs.~${\sim}50M$) compared to our baselines. 
Beyond quantitative gains, our results reveal that the right 2D projections, when aligned with human-centric geometry, offer a powerful and scalable foundation for physically grounded 3D motion synthesis.

%% file: sec/2_relatedwork.tex
\section{Related Work} \label{sec:related_work}
\paragraph{Text-Conditioned Human Motion Synthesis.}
Generating 3D human motion directly from natural language has emerged as a key research direction, offering intuitive control for animation, virtual reality, and robotics.~\cite{sahili2025text}.
Progress in this domain has been driven by large and diverse datasets such as HumanML3D~\cite{HumanML3D}, KIT-ML~\cite{KIT-ML}, BABEL~\cite{BABEL}, and HuMMan~\cite{cai2022humman}, which provide paired text–motion annotations that enable robust language–motion alignment.
Early works employed VAEs~\cite{petrovich2021action, guo2022generating, petrovich2022temos} and GANs~\cite{Ghosh_2021_ICCV} to align textual and motion embeddings, but these models often lacked temporal coherence and expressive diversity. Subsequent diffusion-based approaches~\cite{tevet2022human, zhang2022motiondiffuse, dabral2022mofusion, kim2023flame} reframed motion synthesis as a denoising process conditioned on text, achieving high realism at the expense of computational cost. 
\par
In parallel, a second line of work discretizes continuous motion into symbolic units, enabling transformer-based sequence modeling akin to language generation.  
TM2T~\cite{guo2022tm2t} and T2M-GPT ~\cite{zhang2023generating} first demonstrated that a VQ-VAE + GPT architecture could learn a motion token vocabulary and autoregressively predict motion from text, achieving long-horizon coherence and improved text alignment. MotionGPT~\cite{jiang2023motiongpt} further unified text and motion understanding within a single pretrained model, while MoMask~\cite{guo2024momask} introduced residual vector quantization to reduce reconstruction artifacts and enhance smoothness. Recent works leverage 
LLMs~\cite{touvron2023llama, radford2018improving, team2023gemini} and pre-trained tokenizers~\cite{raffel2020exploring, radford2021learning} for stronger semantic grounding and cross-modal reasoning, improving the interpretation of nuanced textual intent.
\par
Despite these advances, most text-conditioned models operate on scene-agnostic motion data, ignoring physical context and object geometry. Consequently, generated sequences often violate contact constraints or spatial plausibility when placed in real 3D environments.
Our work extends this tokenized text-to-motion paradigm into the realm of text-conditioned scene-aware motion synthesis.
We introduce geometry-grounded tokenization, where discrete motion codes are learned in the presence of local scene geometry, bridging the gap between linguistic intent and physical interaction.

\paragraph{HSI in Motion Synthesis.}
The key design challenge in generating 3D human motion that is physically plausible and contextually appropriate within a 3D environment lies in scene representation and integration, and requires models to reason explicitly about 3D geometry, object affordances, and potential contacts~\cite{sui2025surveyhumaninteractionmotion}. Early approaches such as PROX~\cite{PROX:2019}, PLACE~\cite{zhang2020place} and HULC~\cite{shimada2022hulc} used optimization-based human–scene fitting, leveraging signed distance fields (SDFs) or voxelized occupancy maps or point clouds to ensure contact consistency. 
More recent learning-based models use richer 3D representations. SceneDiffuser~\cite{huang2023diffusion} conditions a diffusion model on 3D point clouds, capturing spatial occupancy and affordance cues; DIMOS~\cite{Zhao2023dimos} employs a 3D diffusion network with explicit contact and penetration losses on voxelized scenes. 
Next, TRUMANS \cite{jiang2024scaling} scales interaction modeling to hundreds of indoor environments using transformer architectures on voxelized scene embeddings. Similarly, TeSMo~\cite{yi2024tesmo} uses 2D semantic maps as a mid-level proxy for 3D layouts, while LINGO~\cite{jiang2024lingo} and UniHSI~\cite{xiao2024unified} introduce affordance and object-centric features extracted from mesh surfaces or language-conditioned priors to improve contextual reasoning. 
LaserHuman~\cite{cong2024laserhuman} introduces a large-scale language-guided scene-motion dataset that couples natural language and 3D dynamic scenes captured in both indoor and outdoor settings, offering multi-person interaction data for joint language-scene grounding. TokenHSI ~\cite{pan2025tokenhsi} represents motion-scene relationships through tokenized spatial embeddings, encoding scene geometry and affordance context as discrete tokens aligned with motion sequences, effectively bridging geometric and linguistic reasoning. 
\par
These trends highlight a growing effort to couple semantic reasoning with physical realism, mostly through increasingly complex 3D representations. CLOPS~\cite{diomataris2025moving}, on the other hand, uses egocentric 2D RGB-D and segmentation images as the sole scene representation, focusing on navigational control from first-person viewpoints rather than language-driven interaction synthesis involving full-body dynamics.
In contrast, our \method~is a lightweight 2D disentangled framework that separates semantic planning from geometric grounding. By using global BEV images for spatial reasoning and local 2D heightmaps for contact fidelity, \method~demonstrates that carefully designed 2D projections of a scene can capture sufficient affordance and geometry information for high-quality motion synthesis. 

%% file: sec/3_method.tex
\section{Method}
\label{sec:method}

\begin{figure*}[t]
    \includegraphics[width=\textwidth]{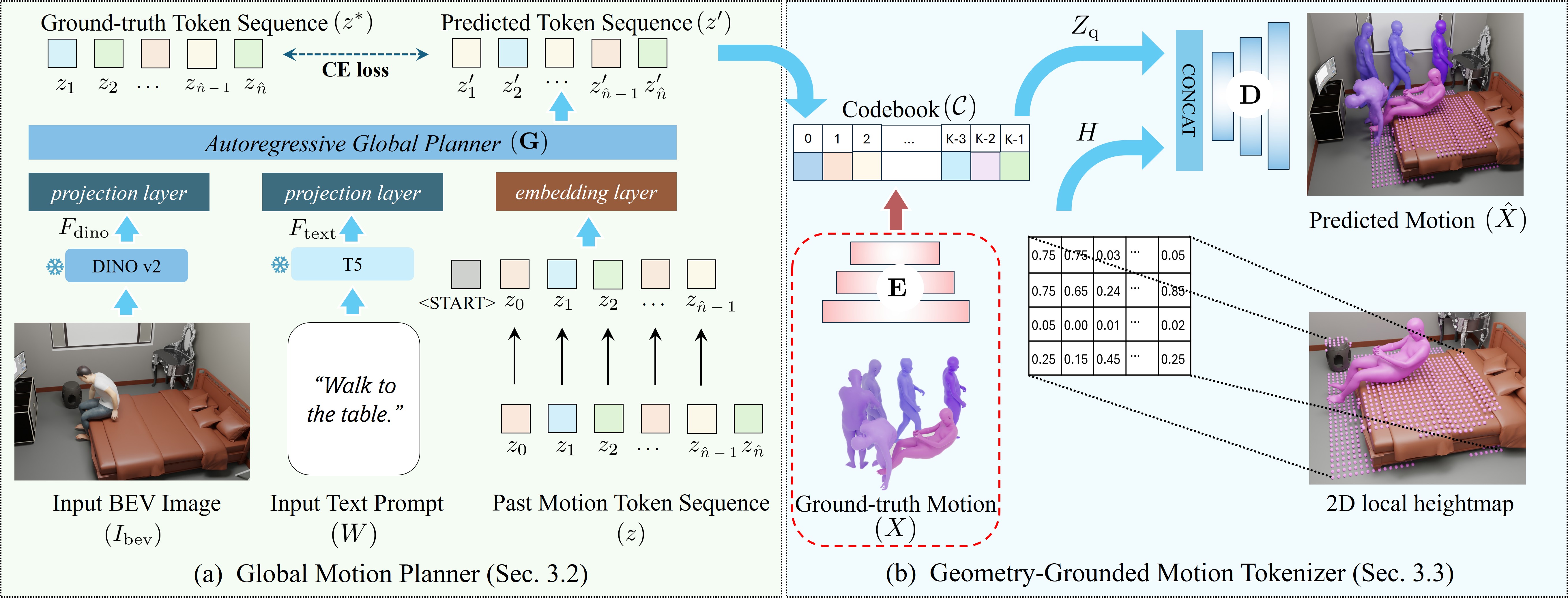}
    \caption{\textbf{Overview of the \method~framework.} 
    \method~disentangles text-conditioned scene-aware human motion synthesis into two stages. (a) The global motion planner predicts discrete motion tokens from text input and DINOv2 scene features extracted from a BEV image. (b) Our geometry-grounded motion tokenizer learns a scene-aware motion vocabulary for mapping these discrete tokens to a continuous 3D human motion. We use 2D local heightmaps around poses to condition our interaction decoder (top right) for fine-grained interaction generation. The red dotted line implies used only during training. Blue arrows follow through the inference pipeline.
    }
    \label{fig:method}
    \vspace{-15pt}
\end{figure*}

\cref{fig:method} depicts how \method~ disentangles global planning from local execution using lightweight 2D scene cues. We now detail our global motion planner framework in \cref{subsec:GPT}, and our geometry-grounded motion tokenizer in \cref{subsec:vqvae}.

\subsection{Problem Formulation}
\label{subsec:formulation}

\textbf{Human Motion Representation.}
We represent the human motion $X$ as a sequence of $N$ poses $X = \{x_i\}_{i=1}^N$. Following standard practices~\cite{HumanML3D, zhang2022motiondiffuse}, we adopt the SMPL-X~\cite{SMPL-X:2019} body structure with $J$ body joints, where each pose $x_i \in \mathbb{R}^D$ is defined as 
\begin{equation}
    x_i = \bracks{
        t_{\delta}, j_{\text{r}}, j_{\text{p}}, j_{\text{v}}, c_{\text{f}}}_i,
\end{equation}
where $t_{\delta} \in \mathbb{R}^3$ is the root translation offset, $j_{\text{r}} \in \mathbb{R}^{J\times 6}$ are the joint orientations in 6D rotation~\cite{Zhou_2019_CVPR}, $j_{\text{p}} \in \mathbb{R}^{(J-1) \times 3}$ are the root-invariant local joint offsets, $j_{\text{v}} \in \mathbb{R}^{J \times 3}$ are the joint velocities, and $c_{\text{f}} \in \mathbb{R}^{4}$ are the binary foot-ground contact indicators. Such over-parameterized representation preserves temporal continuity and contact semantics critical for physically grounded motion synthesis, as shown in many text-based motion synthesis models~\cite{zhang2022motiondiffuse, guo2024momask, guo2022tm2t}. 
\\
\noindent \textbf{Text Prompt Representation.}
Given a natural-language instruction $W$, we obtain its embedding  $F_{\text{text}}$ using a pre-trained T5 tokenizer~\cite{raffel2020exploring}. This feature captures the semantic intent guiding motion planning (\textit{e.g.}, ``walk to the couch'').
\\
\noindent \textbf{Scene Representation.}
To disentangle global planning from local execution, we represent the scene $S$ with two complementary 2D modalities.
To generate high-level global motion, we leverage the representation power of DINOv2 features~\cite{oquab2023dinov2} to provide the vital scene context.
Concretely, we render a single bird's-eye-view (BEV) RGB image $I_{\text{bev}}$ of the scene with the camera positioned at an elevated corner, oriented toward the character's start position.
We then extract a sequence of DINOv2 patch features $F_{\text{dino}}$.
We hypothesize that this representation captures the scene's spatial layout (which areas are walkable) and the semantic location of major objects (\textit{e.g.}, ``couch," ``table") relative to the character's start position.
DINOv2 features come with the added benefit of being efficient to compute, in stark contrast to the large voxel-grids of proxy geometric models used in the existing literature~\cite{wang2022humanise, neural_state_machines}.
\par
For fine-grained physical interactions, we switch to a more precise geometric representation and extract a 2D heightmap $H \in \mathbb{R}^{g \times g}$ of gridsize $g$ centered on the root joint at frame $i$. 
The heightmap covers a predefined area of influence around the character's root joint and is oriented according to the forward direction of the body, ensuring consistent alignment across frames.
This scene representations allow for a causal, autoregressive network design as they are computed from the scene state at the end of the \textit{previous} token during generation.
Overall, given $(F_{\text{text}}, F_{\text{dino}}, H)$  the model learns a mapping
\begin{equation}
\mathcal{G}: (F_{\text{text}}, F_{\text{dino}}, H) \rightarrow X    
\end{equation}
that produces semantically coherent and physically consistent motion.

\subsection{Global Motion Planner}
\label{subsec:GPT}
Our global motion planner serves as the high-level reasoning module that predicts a semantically and spatially coherent motion sequence given a text instruction and a global scene layout. 
To that end, we model long-horizon motion generation as an autoregressive \textit{token prediction} problem in a discrete motion space. 
We train a transformer $\mathbf{G}$
that operates on compact visual and textual cues: the extracted DINOv2 features $F_{\text{dino}}$ and a text embedding $F_{\text{text}}$, to generate a sequence of geometry-aware motion tokens $\{z_i\}^{\hat{n}}_{i=1}$.
Each token $z_i$ within a sequence of $\hat{n}$ tokens indexes a locally consistent motion primitive learned by our geometry-grounded tokenizer (\cref{subsec:vqvae}).
Formally, the planner models the conditional distribution
$P(z_i | z_{<i}, F_{\text{text}}, F_{\text{dino}})$, at each time step $i$.
We use a causal transformer decoder with $L$ layers of multi-head self-attention and feed-forward blocks. Each input token is represented as the concatenation of its learned embedding, a positional encoding, and linear projections of the conditioning vectors $F_{\text{text}}$ and $F_{\text{dino}}$, as shown in~\cref{fig:method} (left). 
At each step, the planner iteratively samples from the categorical distribution
\begin{equation}
z'_i = \argmax_{z_i} P(z_i | z_{<i}, F_{\text{text}}, F_{\text{dino}}),
\end{equation}
thereby constructing a high-level motion plan that respects both semantic intent and scene layout.
\paragraph{Training Objectives.}
We use a cross-entropy loss to maximize the log-likelihood of the ground-truth tokens:
\begin{equation}
\mathcal{L}_{plan} = - \sum_{i=1}^{\hat{n}} \log P(z_i = z_i^* | Z_{<i}, F_{\text{text}}, F_{\text{dino}}),
\end{equation}
where $z_i^*$ denotes the true token index.
To enhance conditioning robustness, we apply classifier-free guidance (CFG)\cite{chang2022maskgit} during training by randomly dropping $F_{\text{text}}$ and $F_{\text{dino}}$ with a fixed probability, allowing adjustable conditioning strength during inference.

\subsection{Geometry-Grounded Motion Tokenizer}
\label{subsec:vqvae}
To map the discrete token sequence $z_i$, used by the global planner, to a continuous motion sequence, $X$, our model needs to learn a discrete motion vocabulary. We train a conditional VQ-VAE to get codebook $\mathcal{C} = \{c_k\}_{k=1}^K$, with a codebook size $K$, where each entry of $\mathcal{C}$ encodes a locally consistent motion primitive.
As illustrated in \cref{fig:method} (right), our VQ-VAE consists of a motion encoder $\mathbf{E}$ and an interaction decoder $\mathbf{D}$, created with 1D temporal convolutional blocks, residual blocks~\cite{he2016deep} and SiLU\cite{silu}.
The encoder network maps a ground-truth motion sequence $X \in \mathbb{R}^{N \times D}$ to a downsampled latent sequence $Z = \mathbf{E}(X)$, where $Z \in \mathbb{R}^{\hat{n} \times D_\text{z}}$. $\hat{n}$ is the number of tokens corresponding to $N$ frames of motion, and $D_\text{z}$ denotes the spatial dimensionality of the latent. Each latent vector $z_i$ in $Z$ is then quantized to its nearest neighbor in the codebook $\mathcal{C}$ to produce $Z_{\text{q}} = \mathbf{Q}(Z)$ ~\cite{van2017neural}.

However, unlike a standard motion VQ-VAE, we propose to explicitly condition our interaction decoder, $\mathbf{D}$, on the local scene geometry. 
We design it to reconstruct the original motion $\hat{X}$ using both the quantized latent $Z_{\text{q}}$ as well as the local heightmap $H$ corresponding to the last pose of the previous motion token,
 ensuring causality and consistency with autoregressive inference.
The reconstruction is defined as
\begin{equation}
\hat{X} = \mathbf{D}(Z_{\text{q}}, H).    
\end{equation}

This novel design forces the discrete tokens in $\mathcal{C}$ to capture not only kinematic patterns but also geometry-specific motion behavior, after being decoded back into the continuous motion space. 
For example, instead of a generic ``bend knees'' token, a learned token in our paradigm may encode ``bend knees to make contact with a surface at height h.'' 
During training, the decoder can minimize reconstruction loss only if the selected token represents a physically compatible interaction given $H$, thereby embedding local contact physics directly into the token space.
This process effectively binds motion semantics to scene geometry, producing a geometry-aware motion vocabulary that generalizes across diverse surfaces and layouts.

\paragraph{Training Objectives.}
We train the VQ-VAE end-to-end to minimize a composite loss 
\begin{equation}
    \mathcal{L}_{VQ} = \lambda_{\text{rec}}\mathcal{L}_{\text{rec}} + \beta ||\text{sg}[Z_{\text{q}}] - Z||_2 
    \label{eq:vq_loss}
\end{equation}
where  $\mathcal{L}_{\text{rec}}$ is a motion reconstruction loss on joint positions, orientations, velocities, and foot-contact indicators. The next term is the commitment loss with $\text{sg}[.]$ denoting the stop-gradient operator. We use exponential-moving-average (EMA) updates with a reset mechanism~\cite{zhang2023generating} to prevent code collapse and balance token utilization. $\lambda_{\text{rec}}$ and $\beta$ are scalar weights of the individual loss components.

After training, we can represent each code in the quantized sequence $Z_{\text{q}}$ by its corresponding index $k$ in the codebook $\mathcal{C}$, thus mapping a continuous motion sequence $X$ into a discrete token sequence $z \in \{0, 1,...,K\}^{\hat{n}}$ that is used by our global planner in \cref{subsec:GPT}. 
This geometry-aware tokenization enables \method~ to reason over high-level motion structure through discrete planning while maintaining fine-grained physical realism during decoding.

\begin{figure}[t]
  \centering
   \includegraphics[width=\linewidth]{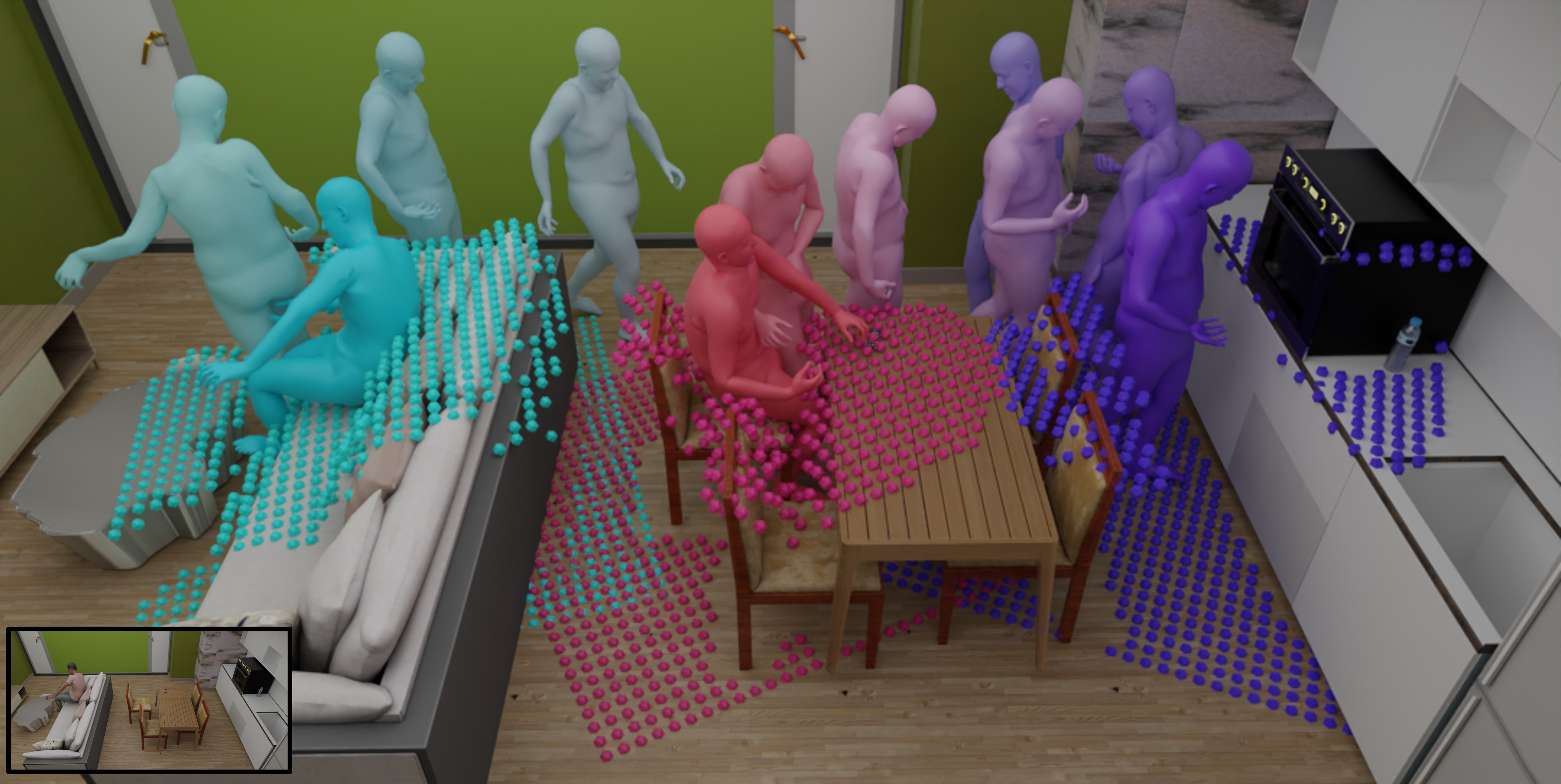}
   \caption{Visualization of long-range motion synthesis in a cluttered indoor environment. \method~ performs geometry-grounded planning by recalculating heightmaps every $t$ frames, enabling globally coherent yet locally feasible motion planning that respects scene geometry. The BEV image input is shown in the inset.}
   \label{fig:result}
   \vspace{-15pt}
\end{figure}
\begin{table*}[t]
    \centering
    \caption{\textbf{Quantitative Evaluation of Motion Generation.} Comparison in terms of scene representation, number of parameters needed for scene encoding, final motion generation quality between baselines, ablated versions, and our full method on the TRUMANS dataset. \textbf{Bold} indicates best. \underline{Underline} indicates second best.}
    \label{tab:quant_baseline}
    \resizebox{\linewidth}{!}{%
        \begin{tabular}{lccccccccc}
        \toprule
        \multirow{2}{*}{Method} &
        \multirow{2}{*}{Scene Representation} &
        \multirow{2}{*}{Scene Encoder} &
        \multirow{2}{*}{\shortstack{\# Trainable \\ Scene Params}} &
        \multirow{2}{*}{FID $\downarrow$} &
        \multirow{2}{*}{Div $\rightarrow$} &
        \multirow{2}{*}{PFC $\downarrow$} &
        \multirow{2}{*}{\shortstack{Cont. \\ $\uparrow$}} &
        \multicolumn{2}{c}{Penetration $\downarrow$} \\
        & & & & & & & & mean & max \\
        \midrule
        Ground truth & - & - & - & - & $2.75$ & $0.24$ & - & - & - \\
        TRUMANS~\cite{jiang2024scaling} & Voxelized 3D occupancy grid & ViT-B Encoder~\cite{dosovitskiy2020image} & $\sim 86M$ & $\underline{0.34}$ & $\mathbf{2.71}$ & $\mathbf{0.58}$ & $\mathbf{0.98}$ & $\underline{1.83}$ & $\underline{11.75}$ \\
        TeSMo~\cite{yi2024tesmo} & 2D floormap + 3D point clouds & ResNet + transformer  & $\sim 35M$ & $0.48$ & $2.02$ & $0.65$ & $0.92$ & $2.11$ & $11.98$ \\
        Humanise~\cite{wang2022humanise} & 3D point clouds & Point Transformer~\cite{zhao2021point} & $\sim 55M$ & $0.82$ & $1.95$ & $1.21$ & $\underline{0.96}$ & $1.95$ & $14.37$ \\
        Scene Diffuser~\cite{huang2023diffusion} & 3D point clouds & Point Transformer~\cite{zhao2021point} & $\sim 55M$ & $0.75$ & $2.03$ & $1.09$ & $0.91$ & $1.71$ & $17.49$ \\
        \midrule
        A5: Single stage transformer & BEV image & DINOv2 (frozen) + linear & $\sim 4M$ & $0.78$ & $1.45$ & $1.67$ & $0.61$ & $3.19$ & $18.99$\\
        A6: CLIP features & BEV image + 2D heightmap & CLIP (frozen) + linear &  $\sim 4M$ & $0.81$ & $2.41$ & $0.87$ & $0.91$ & $1.99$ & $12.89$\\
        A7: W/o trajectory refinement & BEV image + 2D heightmap & DINOv2 (frozen) + linear  & $\sim 4M$ & $0.53$ & $2.01$ & $1.93$ & $0.79$ & $2.78$ & $15.56$\\
        \midrule
        \method~(ours) & BEV image + 2D heightmap & DINOv2 (frozen) + linear & $\sim 4M$ & $\mathbf{0.31}$ & $\underline{2.67}$ & $\underline{0.61}$ & $\mathbf{0.98}$ & $\mathbf{1.81}$ & $\mathbf{11.12}$ \\
        \bottomrule
        \end{tabular}
    }
    \vspace{-10pt}
\end{table*}
\subsection{Inference Loop}
\label{subsec:inference}
At inference, given a text prompt $W$ and a scene $S$, we compute the text features $F_{\text{text}}$ and render the scene context features $F_{\text{dino}}$ using the pre-trained T5 and DINOv2 encoders.
We feed $F_{\text{text}}$ and $F_{\text{dino}}$ as a prefix to our trained planner $\mathbf{G}$, which  autoregressively samples a sequence of motion tokens $z' = \{z'_1, z'_2, ..., z'_{\hat{n}}\}$ with $z'_i \sim P(z'_i | z'_{<i}, F_{\text{text}}, F_{\text{dino}})$ representing a high-level motion primitive consistent with the scene layout and textual intent.
The generated token sequence $z'$ defines a coarse motion plan in the discrete token space. Each token $z'_i$ is then decoded into a continuous motion segment using the proposed interaction decoder $\mathbf{D}$.
\par
To maintain causality, the local heightmap for decoding the current token is re-computed from the updated scene state, centered on the root position of the last frame decoded from the previous token (see \cref{fig:result}).
After each generated motion sequence, we recapture a fresh BEV snapshot of the scene and compute the corresponding heightmap around the character’s new position, allowing the planner to extend long-range trajectories seamlessly across multiple motion cycles.
This recurrent update mechanism ensures that the global plan remains spatially consistent with the evolving scene configuration, enabling smooth and coherent navigation across extended environments.

\paragraph{Trajectory Refinement Module.}
While geometry-conditioned decoding produces physically plausible motion, slight inconsistencies in root trajectory estimation can cause residual artifacts such as foot sliding. To address this, we employ a lightweight trajectory refinement module $\mathbf{R}$ that predicts smoothened root velocities from local joint motion features. 
Specifically, given the local motion features $x^{\text{local}}_i = \bracks{ j_{\text{r}}, j_{\text{p}}, j_{\text{v}}, c_{\text{f}}}_i$, the regressor predicts the root trajectory velocities $ \hat{t}_{\delta, i} = \mathbf{R}(x^{\text{local}}_i)$ using 1D convolutional layers with temporal receptive fields. The network is trained with an L1 reconstruction loss on both absolute root positions and velocities:
\begin{equation}
\mathcal{L}_{\text{traj}} = \lambda_{\text{r}}\norm{t_{\delta} - \hat{t}_{\delta}}_1 + \lambda_{\text{v}} \norm{\Delta t_{\delta} - \Delta \hat{t}_{\delta}}_1.
\end{equation}
At inference, we apply this regressor to the generated motion and replace the root trajectories with refined predictions, effectively reducing foot-sliding artifacts and improving contact consistency. 

%% file: sec/4_experiments.tex
\section{Experiments} \label{sec:experiments}
We next elaborate on our experimental setup, including the dataset, evaluation metrics, baselines and ablations. 
\subsection{Implementation Details}\label{subsec:implementation}
We train the VQ-VAE for approximately $5$K epochs (${\sim}700K$ iterations) on the dataset training set using Adam~\cite{adam} with a base learning rate (LR) of $10^{-5}$ and a batch size of $128$. We use $\lambda_{\text{rec}}=1.0$ and $\beta=0.1$ during training.
We input $N=80$ frames of motion into $\mathbf{E}$ with a downsampling rate of $4$, bringing number of tokens $\hat{n}=20$. For the codebook, we use $K=1{,}024$ codes with dimensionality $D_{z}=512$. The convolution networks use stride 2 with SiLU~\cite{silu} activations and layer normalization in the ResNet blocks.
For the heightmap calculation, we use a grid size of 32 and the grids extend for a distance of $\pm0.6$m around the root position of the character.  

For the autoregressive planning stage, the transformer $G$ consists of $8$ layers with $8$ attention heads each and a hidden state size of $512$. Our input dimensions for $F_{\text{dino}}$ is $768$ with $49$ patches, and $F_{\text{text}}$ is $1024$.
While training $G$, we use a sequence of $20$ tokens and prepend a start-of-sequence token to it. We use a batch size of $32$ on a V100 GPU. It takes about $98$ hours to converge when training the VQ-VAE stage of the model on the training split of the TRUMANS dataset, and $75$ hours for training the planning stage. We use $\lambda_{\text{r}}=\lambda_{\text{v}}=1.0$ for training $\mathbf{R}$.
The full inference pipeline is able to generate a sequence of $80$ motion frames in ${\sim}8$ seconds.
\subsection{Data Preparation and Evaluation Metrics}
\label{subsec:metrics}
We use the TRUMANS dataset~\cite{jiang2024scaling} for training and evaluation. It is a large-scale benchmark for complex Human-Scene Interaction (HSI) containing $15$ hours of human motion data in the SMPLX representation~\cite{SMPL-X:2019} across $100$ distinct scenes. The dataset features diverse interactions, corresponding text annotations, and photorealistic renderings. We adhere to the official data split, using scenes 1-70 (training), 71-80 (validation), and 81-100 (testing).
We provide details of data pre-processing in the Appendix.

\noindent \paragraph{Evaluation Metrics.}
We evaluate the motion tokenizer's reconstruction quality using mean per joint position error (MPJPE) and mean per frame per joint velocity error (MPJVE)~\cite{ghosh2024remos} on the reconstructed motions after token decoding, along with the contact and penetration metrics~\cite{Zhao2023dimos}.
For motion quality, we compute the averaged Fr\'echet Inception Distance (FID)~\cite{heusel2017gans}, which measures the distributional difference between generated and ground-truth movement features. We also calculate latent variance to assess the diversity (Div) of generated movements, and
the physical foot contact score (PFC)~\cite{tseng2022edge} to measure the physical plausibility of the foot movements w.r.t.~the ground plane.
\subsection{Baseline Setup}
\label{subsec:baseline}
We compare \method~ against representative recent methods for human–scene interaction synthesis across different modeling paradigms. 
TRUMANS~\cite{jiang2024scaling} serves as a strong autoregressive diffusion baseline trained on large-scale motion-capture data with voxelized 3D scenes and frame-wise action labels. We use the authors’ official implementation and pretrained weights without modification, as this baseline natively operates on the same dataset.
TeSMo~\cite{yi2024tesmo} represents text-controlled scene-aware motion diffusion models and operates on 2D floor-map inputs with explicit navigation and interaction stages. Since TeSMo was originally trained on synthetic rooms and HumanML3D~\cite{HumanML3D} motions, we finetune its interaction branch on the TRUMANS training split while keeping its text and diffusion modules unchanged.
SceneDiffuser~\cite{huang2023diffusion} represents a diffusion-based model that conditions human motion generation on dense 3D point clouds of scanned scenes. Because the original model was trained on ScanNet~\cite{dai2017scannet}, we evaluate it on TRUMANS dataset using voxelized scene inputs projected to its point-cloud representation for fair evaluation.
Finally, Humanise \cite{wang2022humanise} provides a language-conditioned cVAE baseline trained on synthetic human–scene alignments. We use the released pretrained weights and evaluate them directly on our test scenes for reference, without additional fine-tuning. 
All baselines are evaluated using their original scene representations and conditioning modalities: 3D voxel grids for TRUMANS and SceneDiffuser, and 2D floor maps for TeSMo, while sharing the same test scenes and text prompts.
\begin{table}[t]
    \centering
    \caption{\textbf{Quantitative Evaluation of VQ Reconstruction}. Comparison of motion reconstruction quality after tokenization across ablated versions on the TRUMANS dataset. 
    \textbf{Bold} indicates the best performance. 
    }
    \label{tab:reconstructions}
    \resizebox{\columnwidth}{!}{%
        \begin{tabular}{lccccc}
        \toprule
        \multirow{2}{*}{Variant} &
        \multirow{2}{*}{\shortstack{MPJPE $\downarrow$\\(mm)}} &
        \multirow{2}{*}{\shortstack{MPJVE $\downarrow$\\(mm)}} &
        \multirow{2}{*}{\shortstack{Cont.\\$\uparrow$}} &
        \multicolumn{2}{c}{Penetration $\downarrow$} \\
        & & & & mean & max \\
        \midrule
        A1: Scene-agnostic VQVAE & $25.89$ & $15.98$ & $0.86$ & $4.43$ & $18.91$ \\
        A2a: Heightmap grid (16×16) & $21.48$ & $11.02$ & $0.98$ & $2.12$ & $15.01$ \\
        A2b: Heightmap grid (64×64) & $22.56$ & $10.99$ & $0.97$ & $1.88$ & $\mathbf{14.21}$ \\
        A3: Voxel occupancy grid & $\mathbf{21.32}$ & $12.39$ & $0.92$ & $3.88$ & $16.95$ \\
        A4a: FiLM-style modulation & $27.89$ & $15.66$ & $0.91$ & $2.12$ & $14.89$ \\
        A4b: Cross-attention & $22.86$ & $13.24$ & $0.97$ & $2.25$ & $15.01$ \\
        A7: W/o trajectory refinement & $26.89$ & $15.57$ & $0.79$ & $2.78$ & $15.56$ \\
        \midrule
        Scene aware VQ-VAE~(ours) & $21.88$ & $\mathbf{10.45}$ & $\mathbf{0.99}$ & $\mathbf{1.83}$ & $14.23$\\
        \bottomrule
        \end{tabular}
    }
    \vspace{-10pt}
\end{table}

\subsection{Ablated Versions} \label{subsec:ablations}

To understand the contribution of each design component in \method, we conduct controlled ablations on the TRUMANS test split.
We begin with a minimal scene-agnostic variant and progressively reintroduce the proposed modules to evaluate their influence on motion quality, physical realism, and semantic consistency.
\begin{itemize}[leftmargin=*]
    \bulletitem \textbf{A1: Scene-agnostic VQ-VAE.} We remove the heightmap conditioning from the VQ-VAE decoder, and train the motion tokenizer using only motion data.
    
     \bulletitem \textbf{A2: Heightmap resolution.} We vary the size of the local heightmap grids (16×16 and 64×64) by varying the radius of interaction around the root of the person, to examine the sensitivity of the model to geometric detail in the scene representation.

     \bulletitem \textbf{A3: Voxel occupancy grid representation.} The 2D heightmap is replaced with a 3D voxel occupancy grid~\cite{jiang2024scaling} to test whether volumetric encoding provides additional benefits over lightweight 2D inputs. 
    
    \bulletitem \textbf{A4: Scene-motion fusion strategy.} We evaluate different methods for integrating heightmap feature with motion tokens, including cross attention, FiLM-style modulation~\cite{perez2018film} mechanisms, for the VQ-VAE decoder.
    
    \bulletitem \textbf{A5: Planner-decoder decomposition.} We disable the two-stage setup and train a single transformer that directly predicts full motion sequences, assessing the impact of explicit planning separation.

    \bulletitem \textbf{A6: Scene encoder backbone.} The DINOv2-based BEV encoder is replaced with CLIP~\cite{ramesh2022hierarchical} features to test the effect of semantic scene representations.
    
    \bulletitem \textbf{A7: Trajectory refinement regressor.} We remove the trajectory refinement module to evaluate its contribution to physical consistency.
    
\end{itemize}

%% file: sec/5_results.tex
\section{Results and Discussions}
\label{sec:result}
\begin{figure}[t]
  \centering
   \includegraphics[width=\linewidth]{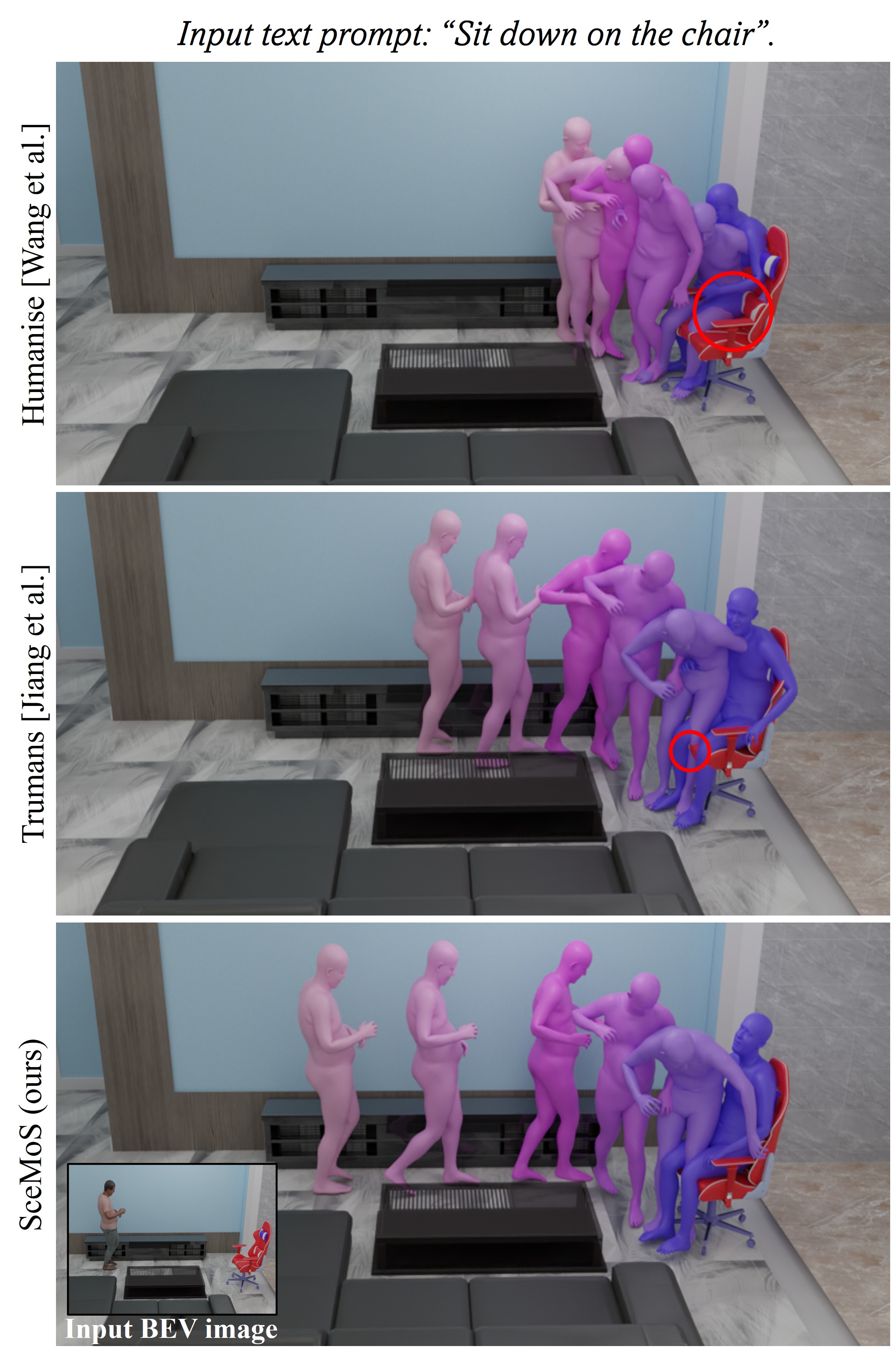}
   \caption{Qualitative Comparison of \method~with recent HSI models. \method~generates motions that are semantically aligned with the input text instructions while maintaining stable contact and smooth transitions. In contrast, we observe some penetrations and misalignment (red circle) in some frames of the baselines.}
   \label{fig:qualitative_comparison}
   \vspace{-15pt}
\end{figure}
We next summarize the results of our experiments and key observations.

\subsection{Quantitative Evaluation}
\label{subsec:quantitative_eval}
\cref{tab:quant_baseline} compares \method~with recent human-scene interaction frameworks in terms of motion realism, diversity, and physical fidelity. The results show that \method~ achieves the lowest FID ($0.31$) and highest contact score ($0.98$) among all evaluated methods, performing on par with the baseline TRUMANS~\cite{jiang2024scaling} that uses voxelized 3D scene data with an order of magnitude higher number of trainable parameters for scene encoding.
Among ablated versions, removing the two-stage setup (A5) significantly degrades fidelity and contact, confirming the benefit of explicit factorization between global planning and local motion decoding. Using CLIP features (A6) instead of DINOv2 for semantic layout also generates poor motion fidelity.
Finally, removing the trajectory refinement (A7) provides modest overall improvements, notably enhancing foot contact plausibility. 
We also report the different scene representations across baseline methods and obtain the scene encoder complexities using standard backbone sizes used in these methods (\textit{e.g.}, Point Transformer\cite{zhao2021point}, ViT-B\cite{dosovitskiy2020image}, ResNet-18\cite{he2016deep}). Across comparable settings, \method~ requires ${<}5M$ trainable scene parameters, which is substantially lower than voxel or point-based encoders.
\par
\cref{tab:reconstructions} reports a quantitative comparison against the ablations of the proposed VQ-VAE under different scene representations, fusion strategies, and refinement settings on the TRUMANS dataset (as discussed in \cref{subsec:ablations}).
Adding scene geometry markedly improves both reconstruction accuracy and physical realism over the scene-agnostic baseline (A1), reducing MPJPE from $25.89$ mm to $21.48$ mm and MPJVE from $15.98$ mm to $10.45$ mm, while raising contact from $0.86$ to $0.98$ and lowering mean/max penetration from $4.4$ / $18.9$ mm to $1.8$ / $14.2$ mm.
This shows that lightweight local geometry cues enable the codebook to capture ground-aware motion and suppress interpenetration.
Across heightmap resolutions (A2), the 32×32 grid achieves the best trade-off. Coarser maps lose elevation detail and denser ones add redundant noise.
Replacing the heightmap with a 3D voxel grid (A3) slightly improves MPJPE but worsens MPJVE and penetration, confirming that dense volumetric inputs are unnecessary for near-surface interactions.
For fusion (A4), simple feature concatenation outperforms FiLM and cross-attention, which destabilize quantization and increase penetration.
Removing the trajectory refinement (A7) further raises velocity and contact errors.
Overall, our scene-aware VQ-VAE yields the lowest spatial and temporal errors and the most stable contact, validating our compact design.
\subsection{Qualitative Results}
 \cref{fig:qualitative_comparison} illustrates qualitative comparisons of \method~ with recent HSI models.
\method~ generates motions that are semantically consistent with the input text prompt (\textit{e.g.}, ``sit down on the chair'') while maintaining stable contact and smooth transitions across surfaces. 
Compared to prior cVAE-based approaches such as Humanise~\cite{wang2022humanise}, our results exhibit tighter spatial grounding and better alignment between the character and supporting surfaces.
Diffusion-based methods such as TRUMANS~\cite{jiang2024scaling} show improved physical plausibility, but produce minor surface penetrations or unstable contact in certain frames.
\par
\cref{fig:result} further illustrates long-horizon motion synthesis using \method~in a complex indoor environment, highlighting the model’s global-local planning synergy. 
As the character transitions between distant functional zones (\textit{e.g.}, sofa $\rightarrow$ table $\rightarrow$ shelf), \method~ continuously recalculates local heightmaps every $t$ frames, allowing the motion to adapt dynamically to changing surface topology while remaining consistent with the global BEV-level plan.
\par
We report additional quantitative results, including task-completion metrics and a detailed user study, in the Appendix.

%% file: sec/6_conclusion.tex
\section{Conclusion}
We presented \method, a framework for text-conditioned scene-aware 3D human motion synthesis that leverages lightweight 2D scene cues, disentangles high-level motion planning from physical execution, and integrates geometric context directly into the tokenized motion space. 
\method~ achieves strong physical realism and contact fidelity on the benchmark TRUMANS dataset, while reducing trainable scene parameters by over an order of magnitude compared to voxel or point-based baselines~\cite{jiang2024scaling, wang2022humanise}. 
Comparison with recent methods and our ablated versions validates our proposed two-stage design and geometry-conditioned tokenization.
Through qualitative results, we show that our method can bridge visual understanding, geometry, and language to generate motions that are both semantically coherent and physically plausible.
\paragraph{Limitations and Future Work.}
 \method~ currently assumes static scenes and is optimized for macro-scale full-body interactions (\textit{e.g.,} sitting, walking). Fine-grained object manipulation (\textit{e.g.,} grasping a cup) remains challenging as our heightmaps are designed to capture large-scale geometry rather than dexterous hand-object affordances. Additionally, our 2D scene cues are tailored for indoor environments; extending this to outdoor scenes with uneven terrain or heavy occlusions would likely require retraining the global planner on diverse outdoor datasets.
 Finally, our inference can still incur moderate latency (${\sim}8s$ per $80$ frames) due to iterative planning and heightmap recomputation. Addressing these factors through adaptive heightmap scaling and dynamic scene perception is a promising future work direction.
 Extending \method~ to dynamic and multi-agent environments is another natural next step toward broader applicability. Advancing along these directions would strengthen \method~ as a general framework for human-scene interaction modeling in more complex, real-world scenarios.

%% file: sec/7_appendix.tex
\section{Appendix}

\setcounter{table}{0}
\renewcommand{\thetable}{A.\arabic{table}}

\setcounter{figure}{0}
\renewcommand{\thefigure}{A.\arabic{figure}}
\subsection{Data Pre-processing}
We train and evaluate our model on the TRUMANS dataset~\cite{jiang2024scaling}. 
As a pre-processing step, we align the human motion data with the corresponding text annotation data, and then downsample the motion data to $20$ fps. 
Following the original data split, we have $\sim 1269$ sequences in the training set, and $~\sim 408$ sequences in the test split. Each sequence varies from $655$ to $1100$ frames depending on the length of motion. Since it is difficult to train on such long sequences, we generate subsequences of $80$ frames by using a sliding window on the full sequence during data loading. We then transform the positions and orientations of the person such that their root joint is at the global origin and the body faces the $Z_+$ direction. The associated scene geometry is transformed accordingly to maintain spatial consistency. We $Z$-normalize all the motion features before feeding them to the network. For the bird’s-eye-view (BEV) scene representation, we use Blender to position a virtual camera at a corner of the room with maximal visibility and render a BEV snapshot of the environment.

\subsection{User Study}
\label{scemos_userstudy}
\begin{figure*}[t]
    \centering
    \includegraphics[width=1.0\textwidth]{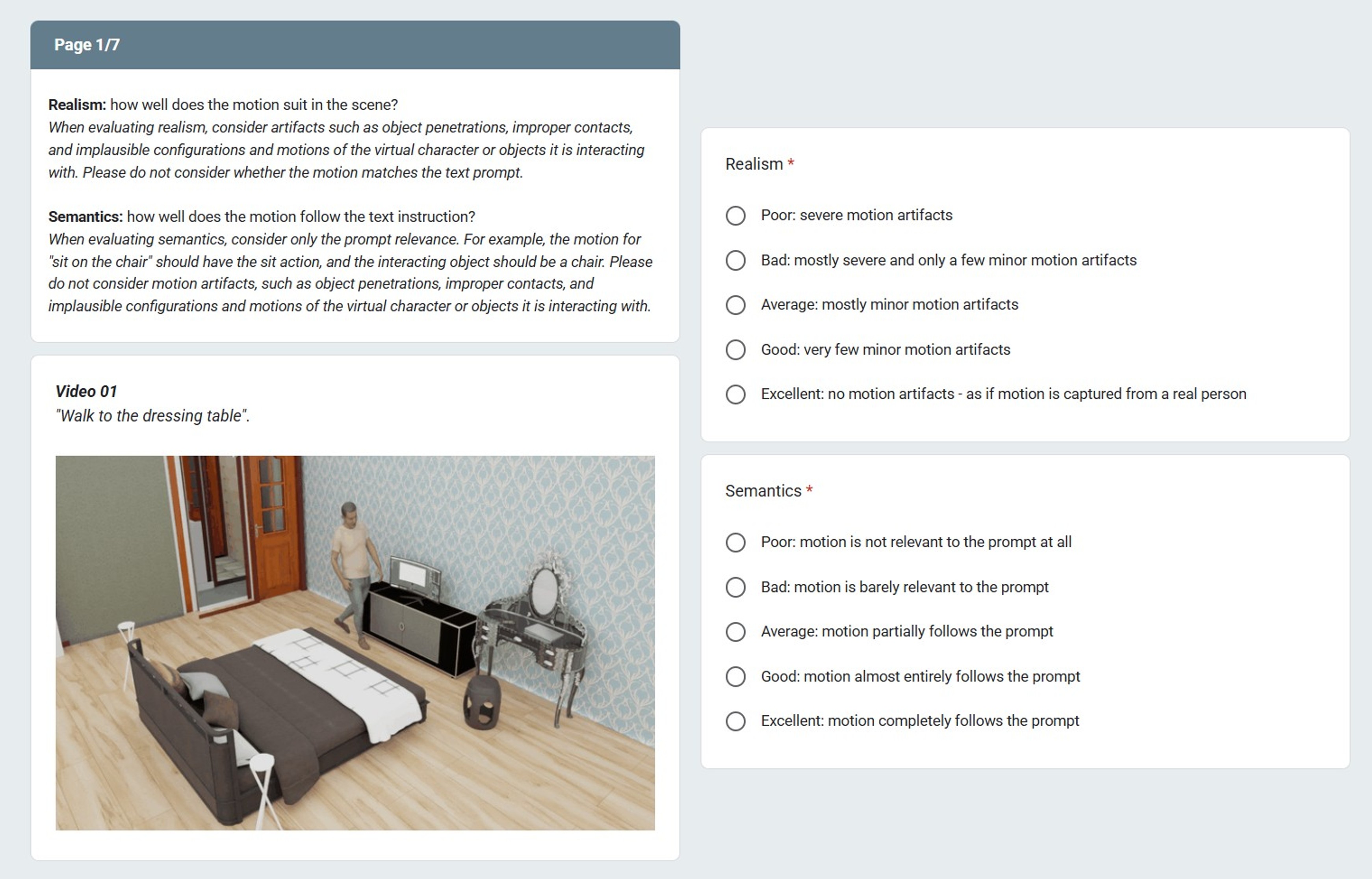}
    \caption{Interface of our user study where we ask participants to rank the motion clips based on `realism' and `semantics', in a 5-point Likert scale.}
    \label{fig:userstudy_interface}
\end{figure*}

\begin{figure*}[t]
    \centering
    \includegraphics[width=0.9\linewidth]{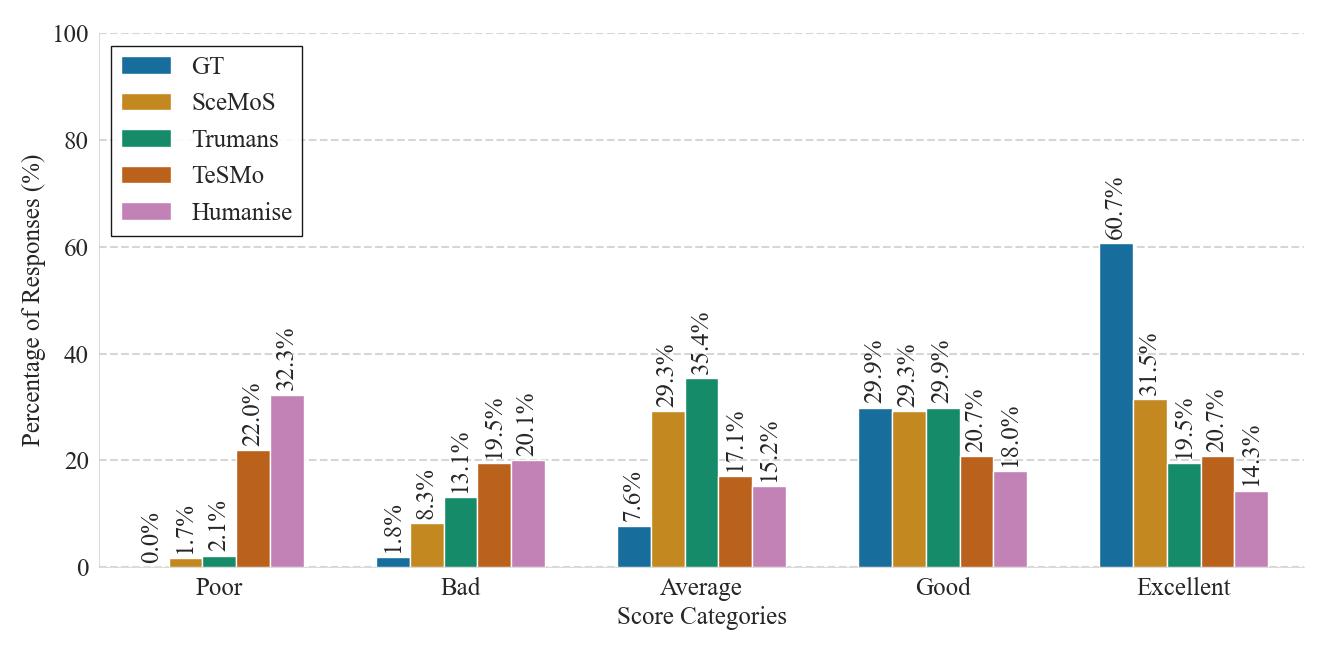}
    \caption{User rating distribution across score categories. \method~ receives the fewest ``Poor / Bad'' ratings and over $60\%$ of its ratings fall into the ``Good / Excellent'' categories.}
    \label{fig:userstudy_distribution}
\end{figure*}

\textbf{Setup.} To assess the visual quality of our results from a human-centered perspective, we conducted a user study. Participants were presented with a sequence of 20 short motion clips generated by different methods (ours, Trumans~\cite{jiang2024scaling}, TeSMo~\cite{yi2024tesmo}, and Humanise~\cite{wang2022humanise}) and rated each clip along two independent axes: Realism and Semantic Alignment. As illustrated in \cref{fig:userstudy_interface}, the interface presents (i) the text prompt, (ii) a rendered visualization of the generated motion interacting with the scene, and (iii) two 5-point Likert scales. `Realism' measures the physical plausibility of the motion within the scene (\textit{i.e.,} presence of foot skating, penetrations, incorrect contacts, or implausible body configurations), while `Semantics' measures how well the motion matches the provided instruction irrespective of physical artifacts.
The five rating categories were “Poor”, “Bad”, “Average”, “Good”, and “Excellent”. 
\\
\textbf{Result.}
We collected responses from 41 participants, drawn from a mix of graduate researchers and industry practitioners. Each participant evaluated all 20 clips, the interface required both ratings to be filled to prevent incomplete submissions. \cref{tab:userstudy} reports the resulting  mean realism and semantic scores across all methods. Ground truth achieves the expected upper bound. Among the generative methods, \method~ yields the best perceptual performance (Realism: 3.41, Semantics: 4.20), outperforming Trumans~\cite{jiang2024scaling}, TeSMo~\cite{yi2024tesmo}, and Humanise ~\cite{wang2022humanise}. Users consistently rated our motions as semantically faithful, even in scenes with challenging geometric layouts. This trend is further supported by the rating distribution shown in ~\cref{fig:userstudy_distribution}.
\method~ receives the fewest ``Poor / Bad'' ratings and over $60\%$ of its ratings fall into the ``Good / Excellent'' categories, indicating strong perceptual stability across different scene contexts.

\begin{table}[ht]
    \centering
    \caption{User study evaluation across all methods. 
    We report the mean $\pm$ standard deviation of participant ratings 
    for \textit{Realism} and \textit{Semantics} in a 5-point Likert scale.}
    \label{tab:userstudy}
    \resizebox{\columnwidth}{!}{
    \begin{tabular}{lccccc}
        \toprule
        Metric & GT & SceMoS & Trumans & TeSMo & Humanise \\
        \midrule
        Realism & $4.36 \pm 0.1$ & $3.41 \pm 0.2$ & $3.38 \pm 0.1$ & $2.19 \pm 0.4$ & $1.83 \pm 0.3$ \\
        Semantics & $4.63 \pm 0.2$ & $4.20 \pm 0.6$ & $3.65 \pm 0.7$ & $3.79 \pm 1.3$ & $3.41 \pm 1.4$ \\
        \bottomrule
    \end{tabular}
    }
    \vspace{-10pt}
\end{table}

\subsection{Additional Quantitative Experiments}
\label{subsec:additional_experiments}
\noindent\textbf{Additional Metrics.}
To also assess how well the motions generated from \method~adhere to semantic intent and spatial targets, we evaluate R-Precision (Top-1)~\cite{guo2022generating} and Goal Accuracy on the TRUMANS test set.
R-Precision evaluates motion-text alignment by measuring the retrieval accuracy of the ground-truth prompt from a set of distractors. Goal Accuracy is defined as the percentage of sequences where the character successfully navigates to within 0.5m of the target object's location. As shown in \cref{tab:additional_metrics}, \method~ outperforms baselines in both metrics, confirming that our geometry-grounded tokens effectively capture both semantic intent and precise spatial targeting.
\\
\begin{table}[h]
    \centering
    \caption{\textbf{Extended quantitative evaluation.} Left: Task-completion metrics on TRUMANS dataset (Goal Accuracy and R-Precision). Right: Generalization performance on the HUMANISE dataset.}
    \label{tab:additional_metrics}
    \resizebox{\columnwidth}{!}{
    \begin{tabular}{lccccc}
        \toprule
        \multirow{2}{*}{Method} &
        \multicolumn{2}{c}{\textbf{TRUMANS}} &
        \multicolumn{3}{c}{\textbf{HUMANISE}} \\
         & Goal Acc. $\uparrow$ & R-Prec. $\uparrow$ & FID $\downarrow$ & Div. $\rightarrow$ & Cont. $\uparrow$\\
        \midrule
        SceMoS    & $\mathbf{0.62}$ & $\mathbf{0.58}$ & $\mathbf{0.95}$ & $1.33$ & $\mathbf{0.74}$\\
        Trumans   & $0.59$ & $0.57$ & $1.02$ & $1.65$ & $0.74$\\
        TeSMo     & $0.49$ & $0.34$ & $1.17$ & $1.21$ & $0.72$\\
        Humanise  & $0.28$ & $0.31$ & $0.99$ & $\mathbf{1.22}$ & $0.72$\\
        \bottomrule
    \end{tabular}
    }
\end{table}

\noindent\textbf{Testing on Additional Datasets.}
To evaluate the generalization capability of our framework, we tested \method~on a subset of the HUMANISE test set~\cite{wang2022humanise} without any architectural changes. We observe competitive performance in terms of diversity and contact metrics (see \cref{tab:additional_metrics}), demonstrating that our 2D scene representation generalizes effectively across different indoor scene distributions.
\\
\noindent\textbf{Backbone Compatibility.}
We also validated that our model is compatible with newer backbones such as DINOv3~\cite{simeoni2025dinov3}. In an initial convergence analysis limited to 500 iterations, we observed that DINOv3 matches the convergence rate of our default DINOv2 encoder, with the training loss dropping from 10.0 to 3.5 ($\pm$ 1.65). This confirms that our framework is compatible with other dense visual backbones and is not strictly tied to the specific DINOv2 architecture.